# PULCINELLA
## A General Tool for Propagating Uncertainty in Valuation Networks

**Alessandro Saffiotti and Elisabeth Umkehrer**

IRIDIA - Université Libre de Bruxelles
Av. F. Roosevelt, 50 - CP 194/6
1050 Bruxelles - Belgium
E-mail: r01507@bbrbfu01.bitnet

**Abstract**

We present PULCinella and its use in comparing uncertainty theories. PULCinella is a general tool for Propagating Uncertainty based on the Local Computation technique of Shafer and Shenoy. It may be specialized to different uncertainty theories: at the moment, Pulcinella can propagate probabilities, belief functions, Boolean values, and possibilities. Moreover, Pulcinella allows the user to easily define his own specializations. To illustrate Pulcinella, we analyze two examples by using each of the four theories above. In the first one, we mainly focus on intrinsic differences between theories. In the second one, we take a knowledge engineer viewpoint, and check the adequacy of each theory to a given problem.

## 1. INTRODUCTION

A new interest has grown up recently in the uncertainty management community. Moving from consideration of efficiency, ease of representation, and generality, a number of techniques for representing and propagating uncertainty in networks have been proposed (e.g. Chatalic et al., 1987; Lauritzen and Spiegelhalter, 1988; Pearl, 1988; Shafer et al., 1987). Moreover, implementations of these techniques have been developed (e.g. Andersen et al., 1989; Hsia and Shenoy, 1989; Zarley et al., 1988; Xu, 1991). However, all the existing systems only propagate uncertainty values according to a single uncertainty theory. This is unfortunate: if we accept that uncertainty theories should be seen as alternatives, rather than rivals (Fox, 1986; Saffiotti, 1987) then we must also accept that for each problem there is a "most adequate" theory, and this theory is in general different from problem to problem. It would be advisable to have a general tool capable of propagating uncertainty according to different uncertainty theories. This would allow us to use the same piece of software for solving different problems that call for different uncertainty management techniques. Moreover, such a tool would be useful for analyzing and comparing different theories in an experimental way.

In the Platonic world of formal theories, a system having the above characteristics exists. Building on their work on belief function propagation (Shafer et al., 1987), Shafer and Shenoy developed a general framework for local computation (Shafer and Shenoy, 1988b) in which the process of network propagation in itself has been abstracted from what is actually propagated. Shafer and Shenoy have shown (Shafer and Shenoy, 1989a) that their framework is capable of modelling both probability and belief function propagation, and that it can capture other existing propagation schemas (i.e. Lauritzen and Spiegelhalter's and Pearl's). This framework has been further generalized by Shenoy (1989), who proposes a class of languages ("valuation-based languages") for building knowledge-based systems. Besides probability and belief functions, other existing uncertainty theories have already been formalized as valuation languages (e.g. Dubois and Prade, 1990).

In this paper, we introduce PULCinella, a tool for Propagating Uncertainty based on the Local Computation technique of Shafer and Shenoy. Pulcinella is a general implementation of valuation based languages, abstracted from a belief function propagation system (Xu, 1991), and it is fully described in (Saffiotti and Umkehrer, 1991a). As such, it may be instantiated to any of the theories which have been formalized as valuation based languages. In particular, four specialization of Pulcinella have already been implemented, namely for propagating probabilities, belief functions, Boolean values, and possibilities. Moreover, Pulcinella makes it easy to implement new theories in it (provided that they can be modelled in the valuation language formalism). Besides describing the tool and the underlying theory, we illustrate the use of Pulcinella for comparing uncertainty theory. In this sense, the AI researcher will find in this paper an analysis of the different results obtained applying different theories to the same problem; and the knowledge engineer will find a discussion of the pros and cons of using different uncertainty theories for modelling the same test-bed problem. Both discussions are based on the results of experiments carried out using Pulcinella.

The rest of this paper is organized as follows. Section 2 reminds some formal background on valuation-based languages and local computations. Section 3 presents Pulcinella. Section 4 shows two full examples of application of Pulcinella. Section 5 discusses these examples and analyzes the differences detected in using different uncertainty theories. Finally, Section 6 concludes.

## 2. THEORETICAL BACKGROUND

### 2.1. VALUATION-BASED LANGUAGES

Shenoy's *valuation-based languages* (Shenoy, 1989) have been abstracted from the axiomatic framework for probabilities and belief functions propagation of Shafer and Shenoy (Shafer, Shenoy and Mellouli, 1987; Shafer and Shenoy, 1988a; Shenoy and Shafer, 1988b). They have been proposed as an alternative to rule-based languages for constructing knowledge-based systems. The language consists of objects, which are used to represent knowledge, and operators, which operate on these objects to make inferences on the knowledge. Two kinds of objects are considered, *variables* and *valuations*, and two operators, *combination* and *marginalization*[1]. We first remind the formal definitions of these elements, and will discuss their interpretation and use later.

*Variables, Frames and Configurations.* We consider a finite set of variables. Each variable may range over a finite set of possible values, called the *frame* for that variable. A *configuration* of a finite non-empty set of variables is an element of the Cartesian product of the frames of the variables in this set.
Denotations: $X$ for the set of variables; $g, h, k$ for subsets of $X$; $W_g$ for the set of configurations of $g$; $x,y$ for single configurations; $a,b,c$ for sets of configurations.

Sometimes we need to project a configuration of one set to another set. A configuration $x$ of $g$ is *projected* to $h$, $g \supset h$, by dropping all the elements in $x$ belonging to $g-h$. It is *extended* to $k$, $k \supset g$, by building the Cartesian product between the configuration and $W_{k-g}$.
Denotations: $x^{\downarrow h}$ for the projection of $x$ to $h$; $x^{\uparrow h}$ for the extension of $x$ to $h$.

*Valuations.* Given a set of variables $h$, we consider a set $V_h$. The elements of $V_h$ are called *valuations* on the set $h$[2]. In our case, valuations are the objects that represent the uncertainty about a set of variables.
Denotations: $V_g$ for the set of valuations on $g$; $V$ for the set of all valuations on subsets of $X$; $G, H$ for single valuations.

*Combination* is any mapping $\otimes: V \times V \to V$, such that, if $G$ and $H$ are valuations on $g$ and $h$, respectively, then $G \otimes H$ is a valuation on $g \cup h$.

*Marginalization.* For each $h \subseteq X$, there is a mapping $\downarrow h: \cup \{V_g \mid h \subseteq g\} \to V_h$, called *marginalization* to $h$, such that, if $G$ is a valuation on $g$ and $h \subseteq g$, then $G^{\downarrow h}$ is a valuation on $h$.

---

[1] A third operator, *solution*, is used for "decoding" the result obtained the propagation. This operator is not relevant to the present discussion, and so it will not be considered.
[2] For the sake of simplicity, we do not take here into account "proper valuations", a subset of the valuations used to restrict the applicability of operators. Thus, the definitions given here are not complete, but they preserve the basic ideas of valuation-based languages.

### 2.2. INTERPRETATIONS

It will be useful to give now some examples of possible interpretations for the syntactical entities of a valuation-based language. This will show in which way a valuation-based language can be used for modelling different existing uncertainty theories. For probability theory, belief-functions, and a Boolean case, the mapping of the theory into the concepts of a valuation-based language have been proposed by Shenoy and Shafer (Shenoy and Shafer, 1988b; Shenoy, 1989). For possibility theory, we use the mapping proposed by Dubois and Prade (1990) building on a previous work by Zadeh (1979).

*Probability:*
Valuations on $h$ are (unnormalized) probability distributions on the configurations of $h$
Combination: If $G$ and $H$ are probability distributions on $g$ and $h$, respectively, then their combination is the probability distribution on $g \cup h$ defined by
$$(G \otimes H)(x) = G(x^{\downarrow g})H(x^{\downarrow h}) \quad \text{for all } x \in W_{g \cup h}.$$
Marginalization: If $h \subseteq g$ and $G$ is a probability distribution on $g$, then the marginal of $G$ for $h$ is the probability distribution on $h$ defined by[3]:
$$G^{\downarrow h}(x) = \Sigma\{G(x,y) \mid y \in W_{g-h}\} \quad \text{for all } x \in W_h$$

*Belief Functions:*
Valuations on $h$ are basic probability assignment (bpa) functions on sets of configurations of $h$.
Combination: If $G, H$ are bpa's on $g, h$ –respectively– then their combination is the bpa on $g \cup h$ defined by[4]
$$(G \otimes H)(c) = \Sigma\{G(a)H(b) \mid (a^{\uparrow(g \cup h)}) \cap (b^{\uparrow(g \cup h)}) = c\}$$
for all $c \subseteq W_{g \cup h}$, $a \subseteq W_g$, $b \subseteq W_h$
Marginalization: If $h \subseteq g$ and $G$ is a bpa on $g$, then the marginal of $G$ for $h$ is the bpa on $h$ defined by
$$G^{\downarrow h}(a) = \Sigma \{G(b) \mid b \subseteq W_g \text{ such that } b^{\downarrow h} = a\}$$
for all $a \subseteq W_h$.

*Boolean:*
Valuations on $h$ are functions $H: W_h \to \{\text{true, false}\}$.
Combination: If $G$ and $H$ are valuations on $g$ and $h$, respectively, then their combination is the valuation on $g \cup h$ defined, for all $x \in W_{g \cup h}$, by
$$(G \otimes H)(x) = \begin{cases} \text{true} & \text{if } G(x^{\downarrow g}) = \text{true and } H(x^{\downarrow h}) = \text{true} \\ \text{false} & \text{otherwise} \end{cases}$$
Marginalization: If $h \subseteq g$ and $G$ is a valuation on $g$, then the marginal of $G$ for $h$ is the valuation on $h$ defined, for all $x \in W_h$, by
$$G^{\downarrow h}(x) = \begin{cases} \text{true} & \text{if there is } y \in W_{g-h} \text{ s.t. } G(x,y) = \text{true} \\ \text{false} & \text{otherwise} \end{cases}$$

---

[3] In all interpretations, we let $G^{\downarrow h}(x) = G(x)$ if $h = g$.
[4] This corresponds to usual (but un-normalized) Dempster's rule of combination (Dempster, 1966).

*Possibility:*
Valuations on h are possibility distributions on sets of configurations of h.
Combination: If G and H are possibility distributions on g and h, respectively, then their combination is the possibility distribution on g∪h defined by

(G⊗H)(x) = min (G($x^{\downarrow g}$), H($x^{\downarrow h}$))  for all x∈ $W_{g \cup h}$.

Marginalization: If h⊆ g and G is a possibility distribution on g, then the marginal of G for h is the possibility distribution on h defined by

$G^{\downarrow h}$(x) = sup {G(x,y) | y∈ $W_{g-h}$}  for all x∈ $W_h$.

### 2.3. REPRESENTING PROBLEMS BY VALUATION SYSTEMS

A set of variables, with their frames, together with a set of valuations, is called a *valuation system*. Intuitively, a valuation system corresponds to a knowledge-base. When we want to use a valuation-based language to solve a problem, we first have to define a valuation system that represents our problem. While doing that, we have to keep in mind the intended intuitive meaning of the syntactical entities of valuation-based languages (which holds independently from the interpretation). To this respect, variables can be seen as representing entities of the domain of discourse, and valuations $V_h$ as representing relational knowledge among the entities represented by the variables in h. The combination operator models aggregation of different fragments of knowledge, and the marginalization operator models a narrowing of the focus of interest by concentrating knowledge on a subset of variables. The following is a pictorial representation of the intended interpretation in terms of real world knowledge of variables and valuations.

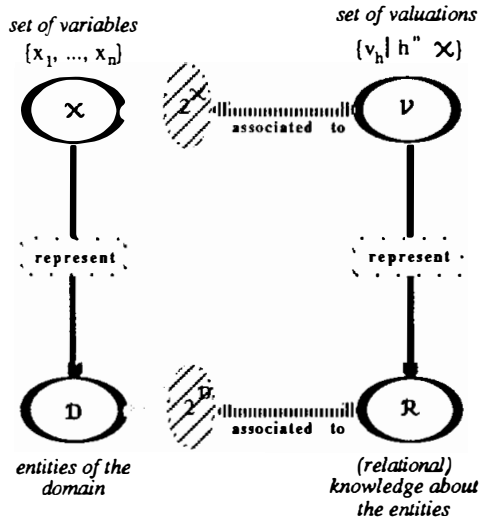

### 2.4. LOCAL COMPUTATION IN VALUATION SYSTEMS

Having defined a valuation system we want now to evaluate it. This means to compute a global valuation on X obtained by combining together all the valuations in our valuation system, and find the marginals of this global valuation to each variable in X. In terms of knowledge, this corresponds to aggregating all the available knowledge together, and to finding the effect of this knowledge on the individual variables. Computing explicitly the global valuation is often unfeasible from the computational viewpoint. However, Shafer and Shenoy (1988b) proposed a general local computation schema for evaluating valuation systems. The computation is local in the sense that combinations of valuations can be performed without extending each valuation to the whole space of the configurations. Shenoy and Shafer have shown that this schema can be applied if the combination and marginalization operators satisfy the following three axioms:

*A1* (Commutativity and associativity of ⊗). Let G, H, K be valuations on g, h and k, respectively. Then:

G⊗H = H⊗G and G⊗(H⊗K) = (G⊗H)⊗K

*A2* (Consonance of ↓). Let G be a valuation on g, and suppose k⊆ h⊆ g. Then:
$(G^{\downarrow h})^{\downarrow k} = G^{\downarrow k}$.

*A3* (Distributivity of ↓ over ⊗). Let G, H be valuations on g and h, respectively. Then
$(G \otimes H)^{\downarrow g} = G \otimes (H^{\downarrow g \cap h})$

All four interpretations described in Section 2.2 satisfy these axioms. Thus, local computations can be used for propagating probabilities, belief functions, Boolean values, and possibilities.

Algorithms based on the above technique have already been proposed and implemented (Zarley et al., 1988; Hsia and Shenoy, 1989; Xu, 1991). These algorithms use a network representation for the valuation system called *Markov Tree*. Mapping a valuation system on a Markov Tree is done in two steps. The first step is to represent the valuation system by a hypergraph: in it, each variable is associated to a node, and each valuation is associated to a hyperedge. The second step is to find a Markov Tree representative of the hypergraph by clustering variables.

### 3.   PULCINELLA

Pulcinella is a system for building and evaluating valuation systems based on Shenoy and Shafer's local computation technique. The system implements the general framework discussed above: it may be specialized to a given uncertainty theory by choosing an interpretation for the objects and the operators. Pulcinella is written in Lisp, and appears as a library of Lisp functions for creating, modifying and evaluating a valuation system. Alternatively, the user can choose to interact with Pulcinella via a graphical interface. The system builds on Hong Xu's implementation of the belief functions propagation technique of Shafer and Shenoy (Xu, 1991). The key move for generalizing Xu's program has been to parametrize it over a set of functions. At the moment, four interpretations of the general framework have been implemented: belief function, probability, Boolean and possibility (in the following, we will use the term "specialization" to refer to an implemented interpretation). By selecting one of these specializations, the user can transform Pulcinella in a probability propagation system, in a belief function propagation system, and so on. Moreover, Pulcinella is meant to be an open system: the set of functions which have to be defined to create a new specialization is very small and with a well defined semantics (reflecting the elements of a valuation-based language). Thus, it is easy for the user to define further specializations. In this Section, we will first show how to use the existing specializations, and then discuss how a new specialization can be created.

### 3.1 USING A PULCINELLA SPECIALIZATION

As far as modelling quantitative knowledge is not concerned, the way to work with Pulcinella is the same for all specializations. We first describe this common part, and then show how qualitative knowledge is modelled in each of the four provided specializations.

The user can model his problem, either graphically or by calling Lisp functions. In terms of Shafer and Shenoy's framework, modelling the problem means to create a valuation system representing his problem. This modelling process comprises two steps. In the first step, the user specifies the structural knowledge of his problem. This means defining all the variables to be used (along with their frames), and indicating which variables are linked together by a relation. By defining the variables and the relations the user implicitly fixes the subsets of variables for which valuations can be specified in the second step. The second step consists in modelling the quantitative knowledge: i.e. in defining the valuations on (some of) the subsets identified by the structural model created. This step depends from the specialization chosen, and will be discussed below. However, during this step, the user should keep in mind that a default valuation is given by the system to each variable and relation if no valuation is defined by the user. Once the valuation system has been completely defined, the user can evaluate it by asking Pulcinella to start propagation. The user can choose if the results must be shown normalized or unnormalized. If the user wants to apply different uncertainty theories on the same problem, he has to specify the structural knowledge only once. He can use the same variables and relations for different specializations.

*Probability Specialization.* In the probability specialization, the user models quantitative knowledge by defining probability distributions for single variables and relations. Notice that this means that a dependency between variables is encoded by a joint probability distribution rather than by conditional probabilities. The default valuation is the uniform probability distribution: if nothing is known about a variable or relation, all configurations are considered to have the same probability. A probability distribution is called normalized if the values attached to the configurations adds up to one.

*Belief-Function Specialization.* In the belief function specialization, the user models the quantitative knowledge by defining basic probability assignment functions on the sets of variables that constitute the structural knowledge. A basic probability assignment function on a set of variables reflects to which extent some subsets of configurations are believed to contain the true configuration, and is expressed by a mapping from subsets of configurations of a set of variables to the interval [0,1]. The value associated to the empty set is always zero. The default valuation is the basic probability assignment function which attach the value 1 to the whole set of configurations. A basic probability assignment function is called normalized if the sum of all its values is 1.

*Boolean Specialization.* The Boolean specialization can be seen as a way to represent categorical knowledge. The quantitative knowledge is modelled by attaching to the configurations of the defined sets of variables either <u>true</u> or <u>false</u>. These values are better understood in term of satisfaction of constraints: a value <u>true</u> for a configuration means that this configuration is acceptable given the constraints of our problem. Accordingly, a relation among variables will be encoded by selecting all those configurations that are admissible, and by associating <u>true</u> to them. The default valuation attaches <u>true</u> to each configuration: if nothing is known about a set of variables, each configuration could be the case. No normalization is defined for Boolean specialization.

*Possibility Specialization.* In possibility theory the user models quantitative knowledge by specifying possibility distributions on the defined sets of variables. A possibility distribution on a set of variables reflects to what extent each configuration of the set is regarded as possible, and is expressed by a mapping from configurations to the [0,1] interval. The default possibility distribution attaches to each configuration the value 1: if nothing is known about a set of variables, all configurations are regarded as completely possible. A possibility distribution is called normalized if at least one element of the frame has possibility value 1.

### 3.2 HOW TO BUILD A NEW SPECIALIZATION

Pulcinella is an open system. The user can build his own specialization with Pulcinella. To build a new specialization the first thing he has to do is to express his theory in terms of the syntactical entities of valuation-based languages, and to prove that the axioms of local computation are satisfied. Having expressed the theory in this way, he may now implement the functions specific to the new specialization. This is made easier by the clear semantics given to these functions: basically, they mirror the concepts and entities of a valuation-based language. These functions may be divided up into three groups:

1. two functions defining the default valuations for variables and relations

2. two functions that implement the combination and marginalization operators.

3. two extra functions for changing the valuations in "some way": one implements the normalization procedure for valuations; the other is called after the propagation has been completed, and allows the builder of a specialization to do some housekeeping before the results are shown to the user.

## 4. EXAMPLES

In this section we will give a couple of examples aimed at illustrating the use of Pulcinella for modelling and solving uncertain problems. We will insist on the difference between modelling the structural knowledge of the problem, and modelling its quantitative knowledge. In each example, the same problem will be formalized in all the four specializations discussed above. However, and interestingly, the structural model remains the same for all of them. These examples will highlight how using different uncertainty theories may require different input and produce different results.

### 4.1. EXAMPLE 1

Our first example is adapted from (Saffiotti, 1987). We want to guess if Francesco will came wearing a black (B), white (W) or polka-dot (P) suite. We do not have any information about Francesco's preference (*state 0*), but we do know that his "Philco" washing machine is out; this makes us believe (say 80%) that he cannot choose W (*state 1*). Later, we remember that Francesco said yesterday that he dislikes mono-chromatic clothes; this is, for the notorious coherence of Francesco, a strong (say 90%) evidence both against B and W (*state 2*).

We first build a structural model for our problem. We define three variables: *Dress*, with frame {B, W, P}; *Philco*, with frame {ok, out}; and *Speech*, with frame {uttered, unuttered}. We then define two relations: *Washing*, between *Philco* and *Dress*; and *Coherence*, between *Speech* and *Dress*. The intended meaning of these elements should be self evident. The following is a graphical representation of our model, as appearing on the screen of Pulcinella:

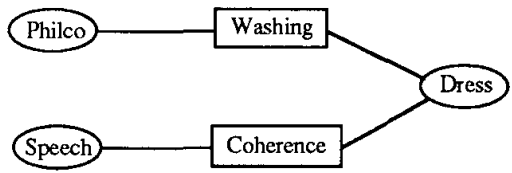

The next step consists in deciding one uncertainty calculus to use, and to specialize Pulcinella to it. Suppose we choose to specialize Pulcinella to probability[5]. We can then enter the (unnormalized) joint probability distributions for our variables and relations. These distributions, which encode the quantitative knowledge in our problem, are shown below[6]:

| $\mathcal{P}_{washing}$ | B | W | P |
|---|---|---|---|
| ok | 1/6 | 1/6 | 1/6 |
| out | 0.2 | 0.1 | 0.2 |

| $\mathcal{P}_{coherence}$ | B | W | P |
|---|---|---|---|
| uttered | 0.025 | 0.025 | 0.45 |
| unuttered | 1/6 | 1/6 | 1/6 |

Notice that we are assuming uniform prior distributions whenever no explicit information is available. Finally, we ask Pulcinella to start propagation. The following table gives the marginal probabilities computed for *Dress* at different moments[7]:

| Value | State 0 | State 1 | State 2 |
|---|---|---|---|
| B | 0.33 | 0.40 | 0.051 |
| W | 0.33 | 0.20 | 0.026 |
| P | 0.33 | 0.40 | 0.923 |

Suppose that we now want to try to use belief functions for modelling our problem. All we need to do, is to specialize Pulcinella to belief functions, and to input the quantitative knowledge of our problem in the form of two basic probability assignments for our relations:

| Subset | $m_{washing}$ |
|---|---|
| ok {B,W,P} | 0.8 |
| out {B,P} | |

| Subset | $m_{coherence}$ |
|---|---|
| uttered {P} | 0.9 |
| unuttered {B,W,P} | |

Intuitively, the subset to which $m_{washing}$ assigns a 0.8 mass represents the fact that the answer to our problem may be any of B, W and P when *Philco* = <u>ok</u>, and any of B and P when *Philco* = <u>out</u>. The remaining 0.2 mass is automatically given to the whole frame. After propagation, we get the following results for the variable *Dress*[8]

| Value | State 0 bel | State 0 pl | State 1 bel | State 1 pl | State 2 bel | State 2 pl |
|---|---|---|---|---|---|---|
| B | 0 | 1 | 0 | 1 | 0 | 0.1 |
| W | 0 | 1 | 0 | 0.2 | 0 | 0.02 |
| P | 0 | 1 | 0 | 1 | 0.9 | 1 |

Next, we consider using possibility theory: we switch Pulcinella accordingly, and enter two possibility distributions for our relations. These distributions, and the one obtained for the variable *Dress* after propagation, are shown below.

| $\Pi_{washing}$ | B | W | P |
|---|---|---|---|
| ok | 1 | 1 | 1 |
| out | 1 | 0.2 | 1 |

| $\Pi_{coherence}$ | B | W | P |
|---|---|---|---|
| uttered | 0.1 | 0.1 | 1 |
| unuttered | 1 | 1 | 1 |

| Value | State 0 | State 1 | State 2 |
|---|---|---|---|
| B | 1 | 1 | 0.1 |
| W | 1 | 0.2 | 0.1 |
| P | 1 | 1 | 1 |

Finally, we consider the case in which we collapse uncertainty to true/false values. The following tables show the values for our relations, and the results obtained, with the Boolean specialization[9].

| $\mathcal{T}_{washing}$ | B | W | P |
|---|---|---|---|
| ok | true | true | true |
| out | true | false | true |

| $\mathcal{T}_{coherence}$ | B | W | P |
|---|---|---|---|
| uttered | false | false | true |
| unuttered | true | true | true |

---

[5] In practice, this reduces to selecting a menu item, or to evaluating the form "(specialize-uncertainty 'probability)".
[6] The valuations for variables are obvious, and will not shown.
[7] States in the table refer to the states in the statement of our story.
[8] For greater readability, we show the results using *bel* and *pl* functions (Shafer, 1976). Remind that pl(A) = 1 - bel(~A).
[9] As noticed above, these values are better understood in term of satisfaction of constraints.

| Value | State 0 | State 1 | State 2 |
|-------|---------|---------|---------|
| B | true | true | false |
| W | true | false | false |
| P | true | true | true |

### 4.2. EXAMPLE 2

The next example has been tailored on an experiment made in modelling the uncertainty present in a problem of fault diagnoses in electricity networks (Gallastegui et al., 1989). For the sake of clarity, both the qualitative and the quantitative knowledge have been greatly simplified. The full experiment, and the actual figures used, are reported in (Saffiotti and Umkehrer, 1991b). We consider here the following fragment of an electricity network:

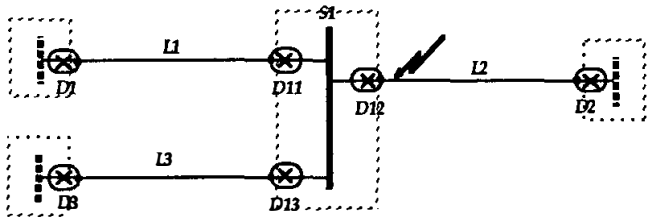

This fragment comprises four "substations", linked by three electricity lines L1, L2 and L3. The substation in the middle includes S1, a big conductive bar used for connecting more lines together. The Di's are "circuit breakers": automatic switches that can isolate two lines when they detect an overload on the part of the network on their "hot" side (marked by a dot in the picture). When an overload is detected, a circuit breaker generate an alarm. There are two kinds of alarm: "instantaneous", for "big" overloads (normally caused by a fault in the line the device is on); or "delayed", for "small" overloads (normally caused by a fault in a neighbour line). All the alarms are sent to the "control room" of some power station. Here, a system engineer is constantly analyzing the incoming alarms to find out what is happening in the network. His goal is to determine if and where there is a fault. We model our electricity network in Pulcinella by the following variables and relations:

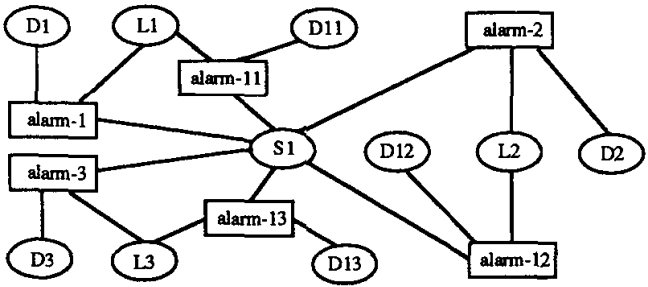

where $Di$'s are variables representing circuit breaker states, with possible values ok (no alarm), del (delayed alarm), and inst (instantaneous alarm); $Li$'s and $S1$ represent line states, with frame {ok, fault}; and the $alarm-i$'s relate generation of alarms by breakers with states of neighbour lines.

The quantitative knowledge given by the experts is not very rich. Essentially, it says that:

1. alarms are not very reliable: in roughly 10% of the cases, they do not correspond to the real situation (alarm generated without fault, or fault occurring without alarm);
2. if an instantaneous alarm is generated (correctly), the fault is 70% of the cases in the line the breaker is on, and 30% in the next one; the reverse holds for delayed alarms.

The following tables show how this knowledge has been coded into the relation alarm-1 by an (unnormalized) joint probability distribution $P$, a possibility distribution $\Pi$, a Boolean constraint $\mathcal{T}$, or a basic probability assignment $m$, respectively. Distributions for the other alarm-$i$'s are similar; while those for the alarm-$ij$'s are different.

| $P(\bullet)$ D1 L1, S1 | ok | del | inst |
|---|---|---|---|
| ok, ok | 0.9 | 0.1 | 0.1 |
| ok, fault | 0.05 | 0.6 | 0.2 |
| fault, ok | 0.05 | 0.2 | 0.6 |
| fault, fault | 0.001 | 0.1 | 0.1 |

| $\Pi(\bullet)$ L1, S1 | ok | del | inst |
|---|---|---|---|
| ok, ok | 1 | 0.1 | 0.1 |
| ok, fault | 0.1 | 0.7 | 0.3 |
| fault, ok | 0.1 | 0.3 | 0.7 |
| fault, fault | 0.1 | 1 | 1 |

| $\mathcal{T}(\bullet)$ | ok | del | inst |
|---|---|---|---|
| ok, ok | true | false | false |
| ok, fault | false | true | true |
| fault, ok | false | true | true |
| fault, fault | false | true | true |

| L1, S1 / D1 | ok, ok | ok, fault | fault, ok | fault, fault | $m(\bullet)$ |
|---|---|---|---|---|---|
| ok del inst | | | | | 0.096 |
| " | | | | | 0.256 |
| | | | | | 0.348 |
| | | | | | 0.016 |
| | | | | | 0.024 |
| | | | | | 0.064 |
| | | | | | 0.096 |

We now imagine a scenario in which a fault occurs on line L2 very near to D12 (see the above picture): as a consequence, a "delayed" alarm is sent by D2, and a "instantaneous" alarm by D12. Moreover, we imagine that D1 has been adjusted incorrectly (it happens), and is too sensitive: thus, also D1 sends a "delayed" alarm.

These three alarms will be received in the control room, and the system engineer will have to find out what has happened. Three main hypotheses are compatible with this set of alarms, shown in order of preference:

1) Fault in L2; alarm from D1 spurious;
2) Fault in S1; missing alarm from D3, and alarm from D12 spurious;
3) Fault in L1; D11 not working (missing alarm & did not open), missing alarm from D3, and alarm from D12 spurious.

The following tables summarize the results obtained for this example with the probability, possibility, Boolean, and belief function specializations of Pulcinella[10]

1. *After receiving the alarm from D2*

| Var | Probab $\mathcal{P}$(fault) | Possibility N(fault) | Possibility $\Pi$(fault) | Belief Functions Bel(fault) | Belief Functions Pl(fault) | Boolean $\mathcal{T}$(fault) | Boolean $\mathcal{T}$(ok) |
|---|---|---|---|---|---|---|---|
| L1 | 0.002 | 0.3 | 1 | 0.02 | 1 | true | true |
| L2 | 0.226 | 0.3 | 1 | 0.02 | 1 | true | true |
| L3 | 0.002 | 0.3 | 1 | 0.02 | 1 | true | true |
| S1 | 0.087 | 0.7 | 1 | 0.16 | 1 | true | true |

2. *After receiving the alarm from D12*

| Var | Probab $\mathcal{P}$(fault) | Possibility N(fault) | Possibility $\Pi$(fault) | Belief Functions Bel(fault) | Belief Functions Pl(fault) | Boolean $\mathcal{T}$(fault) | Boolean $\mathcal{T}$(ok) |
|---|---|---|---|---|---|---|---|
| L1 | 0.002 | 0.3 | 1 | 0.02 | 1 | true | true |
| L2 | 0.972 | 0.9 | 1 | 0.60 | 1 | true | false |
| L3 | 0.002 | 0.3 | 1 | 0.02 | 1 | true | true |
| S1 | 0.013 | 0.7 | 1 | 0.16 | 1 | true | true |

3. *After receiving the alarm from D1*

| Var | Probab $\mathcal{P}$(fault) | Possibility N(fault) | Possibility $\Pi$(fault) | Belief Functions Bel(fault) | Belief Functions Pl(fault) | Boolean $\mathcal{T}$(fault) | Boolean $\mathcal{T}$(ok) |
|---|---|---|---|---|---|---|---|
| L1 | 0.219 | 0.3 | 1 | 0.03 | 1 | true | true |
| L2 | 0.919 | 0.9 | 1 | 0.60 | 1 | true | false |
| L3 | 0.002 | 0.3 | 1 | 0.03 | 1 | true | true |
| S1 | 0.141 | 0.7 | 1 | 0.29 | 1 | true | true |

## 5. DISCUSSION

Beside illustrating the use of Pulcinella for modelling uncertain problems within a given specialization, the examples above show how Pulcinella may provide useful help in comparing different uncertainty management techniques. The advantage of having this capability is twofold. In theoretical research, Pulcinella allows us to play with uncertainty theories, and to track differences between them. In knowledge engineering, it may be used as a tool for choosing, on an experimental basis, the uncertainty treatment technique that better fits our problem and our needs. The following discussion will illustrate both uses[11]. In our examples, we recognize three basic categories of differences:

1) differences in the data we have to provide (which data, how many, in which form, ...);

2) differences in the results due to differences in the data we provided; and

3) differences in the results due to differences in the mechanisms used in the theories.

We first consider Example 1. Concerning category 1 above, the quantitative knowledge stated in the problem has to be coded differently in the four specializations. In the probabilistic case, the need to specify completely the joint distributions $\mathcal{P}_{washing}$ and $\mathcal{P}_{coherent}$ has obliged us to replace in some way the missing information. In particular, in $\mathcal{P}_{washing}$, the 80% probability of (BvW) has been converted into exact probability values for B and P individually (assuming equiprobability); a similar operation has been made for $\mathcal{P}_{coherent}$. On the other hand, in both the belief function and the possibilistic specializations knowledge has been expressed at exactly the level of granularity that is available. However, it must be noticed that using basic probability assignments to encode knowledge may sometimes be less intuitive (mainly because we have to work on subsets). Finally, our knowledge has easily been reduced in the obvious way in the Boolean specialization.

Moving now to the analysis of the results (categories 2 and 3 above), we first notice that —as expected— all four specializations agree from the qualitative viewpoint. The major quantitative differences show up between the results of the probabilistic specialization and those of the belief function (or possibilistic) one. These differences must be tracked back to differences in the data used, mainly because of the additional hypotheses (equiprobability) introduced in the probabilistic case. Remarkably, these differences are particularly evident in those cases (States 0 and 1) in which ignorance is predominant: here, in spite of our ignorance, the probabilistic approach needs (and gives) precise figures[12]. Another interesting difference arises between the belief function and the possibilistic cases. Here, differences in results do not originate from differences in the inputs given, but rather from the different combination mechanisms. To wit, consider State 2: the Philco evidence and the Speech evidence are both denying the W hypothesis. These two items of evidence have been combined into one much stronger evidence against W by Dempster's rule in the belief function case, resulting in a very small (0.02) plausibility for W. On the other hand, the stronger of them has simply been selected through the MIN operator in the possibilistic approach, giving a possibility value of 0.1 for W.

The above considerations illustrate how to use Pulcinella for analyzing the different behaviours of uncertainty treating theories. However, we do not have a pragmatic unit for measuring how much do a given theory fits our needs. Still, Pulcinella may be also used as a tool for evaluating uncertainty formalisms in the light of the uses to which they are to be put. To this respect, we now consider Example 2. As before, we first consider the differences in the inputs required by the different techniques (category 1 above), and than the differences in the output produced (categories 2 and 3). However, we try to take here a more "knowledge engineering" viewpoint.

Probability theory requires several values that are not available in our data. It is common practice to produce missing values by resorting to some symmetry consideration, or by using the principle of maximum entropy. Accordingly, we had to introduce a number of equiprobability hypotheses for our missing values (e.g. in computing $P(\{ok, ok, fault\})$. Also, notice that the amount of subjective estimates required by the probabilistic approach greatly increases the risk of inconsistency among them (but see Duda et al., 1986). On the contrary, the process of supplying data is extremely simple in the possibility and the belief function specializations: the expert must supply just the data she knows, and consign the rest to silence. We do not need to force her (or the figures she supplies) to say something they were not meant to. However, in the belief function case, the translation of this data into basic probability assignments may sometimes be hard. The difficulty of the Boolean case is somewhat complementary to that of the probabilistic case: the data given by the expert has to be rounded very roughly, and she may feel uncomfortable with the approximations obtained. A possible reaction to this rude attitude is to try to split general rules into more specific ones. The aim of this would be to reduce uncertainty by explicitly accounting for the possible exceptions to rules. Though this constitutes a stimulus for the expert that may sometimes result in a better explanation of her knowledge, the strongly empirical, tangled and "artistic" nature of electrical fault diagnostic knowledge discouraged us from using the Boolean specialization for our problem.

---

[10] For better readability, we show the results of the possibilistic case as necessity/possibility pairs. Remind that $N(A) = 1 - \Pi(\sim A)$.

[11] This is not meant to be a general comparative analysis of different UR techniques. Our sole aim here is to show how Pulcinella may provide useful information in this perspective.

[12] «Wovon man nicht sprechen kann, darüber muß man schweigen» (What we cannot speak about we must consign to silence) (Wittgenstein, 1921).

We now switch to analyzing the results. Our four specializations have produced results that are both quantitatively and qualitatively different. The first phenomenon we notice is that the Boolean specialization does not suggest the possibility of a fault in S1. The causes of this reside both in the input given and in the combination mechanism used. As for the input, the knowledge encoded in our relations allows us to infer that the fault is, e.g., either in S1 or in L1. Yet, no relation encodes knowledge which allows us to infer the S1 hypothesis alone. Discrimination between S1 and L1 is, on the other side, captured by differences in the given weights in the other approaches. As for the mechanism, aggregating knowledge by an AND operation does not allow to perform that "counting" of evidence that seems necessary if we want to accumulate items of weak evidence together. In our example, the evidence given by D2 only partly supports the hypothesis S1; however, we would expect further support for S1 coming from D1 to reinforce the S1 hypothesis. One way to fix this problem in the Boolean specialization is to add new rules (e.g. "IF at least 2 among D1, D2 and D3 send a delayed alarm, THEN S1 is faulty"); but the lesson taken from this story seems to be that considering uncertainty in our problem is necessary, if we want to preserve inferential power without having to drown in a see of specialized rules.

The second qualitative difference we want to notice regards the hypothesis "L1 = fault", which is suggested by the possibilistic and the probabilistic specializations, but not by the belief function one. The origin of this is again in the way we have coded our data: a delayed alarm does not support, in the belief function case, a fault in the adjacent line *individually*, but a fault in the adjacent *or* in the next lines[13]. On the contrary, in the possibilistic and the probabilistic cases, the evidence given by the alarm is spread among each hypothesis individually. A related difference concerns the hypothesis "L3 = fault", suggested by the possibilistic specialization only. This "over-inferencing" is rooted in our rather strong attitude when defining the possibility distributions for the *alarm-i*'s relations (viz. the frequential knowledge about the localization of the fault has been converted to a measure of (im)possibility). A less committed interpretation of our data could lead to the following distribution

| $\Pi(\bullet)$ | ok | del | inst |
|---|---|---|---|
| ok, ok | 1 | 0.1 | 0.1 |
| ok, fault | 0.1 | 1 | 1 |
| fault, ok | 0.1 | 1 | 1 |
| fault, fault | 0.1 | 1 | 1 |

where the information about the relative support given by an alarm to a near fault or a far fault (which is not, strictly speaking, a matter of possibility) has been ignored. Using this encoding, we would get a result suggesting L2 alone (like in the Boolean case). The lesson to be taken here seems to be that possibility theory may provide the additional inference power that we need in our problem, but this power cannot be "fine-tuned" easily.

As for the quantitative differences, the most important one (which might be regarded as qualitative as well) concerns the hypothesis "S1 = fault". This hypothesis is reinforced by the arrival of the alarm from D1 in the probabilistic and belief function specializations, but not in the possibilistic one. The cause here is only the combination mechanism used: like the AND of the Boolean case, the MIN operator does not allow us to perform that "counting", which seems necessary in our problem to accumulate evidence correctly. As a consequence, the answers given by the possibilistic specialization to our problem are meaningful mainly from the qualitative viewpoint (they allow us to focus on those hypotheses which are possible), but they are fairly poor from the quantitative one. To this respect, we notice that the results given by the probabilistic specialization are very rich from the quantitative viewpoint; though, because of our straining the input data, this precision may be somehow unjustified. The belief function specialization seems to provide the trade-off between precision and non-commitment that best fits our knowledge. Yet, the computational complexity inherent in belief functions shows up in running the full experiment: the final choice of the uncertainty treating technique to be used for solving the full scale diagnostic problem will have to take this factor into consideration.

## 6. CONCLUSIONS

We have presented PULCinella (Propagating Uncertainty using Local Computation), and illustrated its use as a comparison tool. The key for Pulcinella's comparison power lies in the separation made between the process of modelling the structural knowledge of a problem, and that of modelling its qualitative knowledge. Once a structural model has been decided, we can superimpose any of the available uncertainty calculi on it. There are at least two places where Pulcinella is expected to be useful. First, on the desk of the theorist involved in the comparative study of uncertain reasoning models. For her, Pulcinella might play the role of a slide-rule when empirically testing the behaviour of uncertainty theories over sample problems of academic interest. The fact that the structural model of the problem is the same for all theories ensures the soundness of the test. Second, on the desk of the knowledge engineer. Here, Pulcinella might prove helpful in checking out different uncertainty management techniques for solving the problem at hands, and then judging—on this experimental basis—which one appears to be the most adequate to our case. In particular, we may test, for each theory, both the input requirements and the results, and evaluate how the available data is accommodated for, or our expectations satisfied.

---

[13] But notice that, differently from the Boolean case, further evidence may convert this support to one for the adjacent line individually.

### Acknowledgements

The work reported here has greatly benefit from discussions with (and comments from) Yen-Teh Hsia, Robert Kennes, Bruno Marchal, and Philippe Smets. Hong Xu provided invaluable assistance during the development of Pulcinella. Inaki Laresgoiti and Francesco Ferri have been worthy sources of inspiration for our examples. Work by the first author has been partially supported by the ARCHON project, funded by the Commission of the European Communities under the ESPRIT-II Program, P-2256. Work by the second author has been supported by a grant of IRIDIA, Université Libre de Bruxelles.